\documentclass{article}
\usepackage{graphicx} 
\usepackage{xcolor}
\usepackage[margin=0.72in]{geometry}
\usepackage{multirow}
\usepackage{amsmath}
\usepackage{amsfonts}
\usepackage{hyperref}
\usepackage{color}
\usepackage{array}
\usepackage{bm}
\usepackage{authblk}
\usepackage{float}

\usepackage[parfill]{parskip}

\providecommand{\norm}[1]{\left\Vert#1\right\Vert}

\title{A Multimodal PDE Foundation Model for Prediction and Scientific Text Descriptions}

\author[1]{Elisa Negrini\thanks{Corresponding author: \href{mailto:enegrini@math.ucla.edu}{enegrini@math.ucla.edu}}}
\author[1]{Yuxuan Liu}
\author[2]{Liu Yang}
\author[1]{Stanley Osher}
\author[1]{Hayden Schaeffer}

\affil[1]{Department of Mathematics, University of California Los Angeles, Los Angeles, CA}
\affil[2]{Department of Mathematics, National University of Singapore, Singapore}

\date{}

\begin{document}

\maketitle
\begin{abstract}
    Neural networks are one tool for approximating non-linear differential equations used in scientific computing tasks such as surrogate modeling, real-time predictions, and optimal control. PDE foundation models utilize neural networks to train approximations to multiple differential equations simultaneously and are thus a general purpose solver that can be adapted to downstream tasks.  Current PDE foundation models focus on either learning general solution operators and/or the governing system of equations, and thus only handle numerical or symbolic modalities. However, real-world applications may require more flexible data modalities, e.g. text analysis or descriptive outputs. To address this gap, we propose a novel multimodal deep learning approach that leverages a transformer-based architecture to approximate solution operators for a wide variety of ODEs and PDEs. Our method integrates numerical inputs, such as equation parameters and initial conditions, with text descriptions of physical processes or system dynamics. This enables our model to handle settings where symbolic representations may be incomplete or unavailable. In addition to providing accurate numerical predictions, our approach generates interpretable scientific text descriptions, offering deeper insights into the underlying dynamics and solution properties. The numerical experiments show that our model provides accurate solutions for in-distribution data (with average relative error less than 3.3\%) and out-of-distribution data (average relative error less than 7.8\%) together with precise text descriptions (with correct descriptions generated 100\% of times). In certain tests, the model is also shown to be capable of extrapolating solutions in time. 
\end{abstract}

\section{Introduction}
\let\thefootnote\relax\footnotetext{Code for this work is available at \url{https://github.com/enegrini/MOL-LLM.git}.}
Neural networks have become increasingly important for solving non-linear differential equations, with applications in climate modeling, financial forecasting, biological systems analysis, and structural optimization (see for instance \cite{poznyak2019survey, sezer2020financial, de2023machine}). Their ability to model complex, non-linear relationships allows for efficient and accurate predictions for various scientific computing tasks such as surrogate modeling, real-time predictions, and optimal control. Previous work in deep learning for partial differential equations (PDE) have focused on learning either the solution operator, which maps input functions to their solutions, or the governing system of equations, which describes the constitutive model based on observations of state variables \cite{chen1995universal, lin2021accelerated, li2020fourier, lu2021learning,zhang2024bayesian, schaeffer2017learning}. These approaches, however, tackle one task at a time and are limited to the use of numerical data.

Building on the observation that families of differential equations frequently share fundamental characteristics, recent work has introduced transformer-based architectures to enable simultaneous encoding of various parametric differential equations \cite{yang2023context, yang2023fine, yang2024pde, cao2024vicon, liu2024prose,sun2024towards,liu2024prosefd,jollie2024time}. Although effective, these methods require structured input and output data, with vanilla ICON~\cite{yang2023context} focusing on numerical data and PROSE utilizing numerical and symbolic data \cite{liu2024prose,sun2024towards}.
In the work, we consider additional modalities as both inputs and outputs to the model. Often, one has access or would like to produce heuristic descriptions of the observed dynamics that are neither in symbolic nor numerical form, but instead come as textual descriptions. For example, consider modeling the dynamics of a complex ecological system: the numerical inputs can include measured population levels, while textual inputs could describe key processes such as predator-prey interactions or migration patterns. Similarly, in material science, numerical data could represent experimental results, while textual inputs provide the governing equation or describe the experimental setup. By combining these modalities, the model may better capture the underlying rules and provide more accurate and contextually informed predictions. The utilize of mathematical formulae, text descriptions, and numerical values can provide a more robust development toward a PDE foundation model. Note that in ICON-LM~\cite{yang2023fine} both textual and numerical prompts were provided as inputs; however, the model does not generate text descriptions since the outputs of ICON-LM are the numerical predictions. 

Consider the following parametrized differential equation:
\begin{align}\label{system}
    \begin{cases}
        \mathcal{F}(u(x,t;c))=0 \quad (x,t)\in \Omega\times[0,T]\\
        \mathcal{B}(u(x,t;c))=0 \quad (x,t)\in \partial\Omega\times[0,T]\\
        u(x,0;c)=\mathcal{G}(x;c), \quad x\in \Omega,; c \sim \mathcal{D}
    \end{cases}
\end{align}
where $\mathcal{F}$ denotes the governing equation, $\mathcal{B}$ denotes boundary conditions. The initial condition $\mathcal{G}$ is a generating function, and $c$ denotes the parameters that determine the initial conditions from distribution $\mathcal{D}$. The objective is to develop a single neural network model capable of approximating numerically and describing in text the solution operators for a range of governing equations $\mathcal{F}$ in \eqref{system}, which can include ODEs and PDEs. We will assume periodic boundary conditions for uniformity in experiments.

The text input may contain either the equation to solve or a textual description of the system. This is important in applications where one may only have a partial model or a description of the underlying process. The numerical data includes the equation's parameters and initial conditions. For instance, suppose we want to solve the heat equation with parameter $c = 0.003$ and initial condition $u(x,0) = u_0(x)$. The input to our model could be:
\begin{itemize}
    \item A formal description: \emph{``The given equation is $u_t = c u_{xx}$ where $c = 0.003$ and $u(x,0) = u_0(x)$."}
    \item A more general description: \emph{``Solve the heat equation with parameter 0.003 and initial condition $u_0(x)$."}
\end{itemize}
In our experiments, we primarily used the first type of input, where the equation is explicitly provided as a formula. However, given sufficient training examples, we expect the model to also perform well when only a descriptive name or general description of the equation is provided.
With the input \emph{``The given equation is $u_t = c u_{xx}$ where $c = 0.003$ and $u(x,0) = u_0(x)$,"} our model outputs:
\begin{enumerate}
    \item \textbf{Numerical Predictions}: The solution to the PDE at user-defined query locations.
    \item \textbf{Scientific Text Descriptions}: A sentence or more that describes the scientific properties of the equation or the solution. For instance:
    \begin{itemize}
        \item Physical process modeled by the equation: \emph{``The heat equation is a parabolic PDE that models the spread of heat in a material over time."}
        \item Alternative numerical methods to approximate the solution: \emph{``The heat equation can be numerically solved using a combination of forward-time and central-space finite difference methods."}
    \end{itemize}
\end{enumerate}
The generality of the generated text descriptions depends on the training set. For instance, descriptions of other properties of interest may be included in the training set, such as the long-term behavior of solutions, the presence of shocks or rarefactions, etc. We refer to Section \ref{sec:text_gen}, Table \ref{tab:description_comparison} for various examples of generated text. The text is collaboratively trained in the neural network with the goal of generalizing text descriptions to new systems. 

As a representative example of our model's capabilities, in Section \ref{sec:results} we show that our model can determine whether the solution to a given conservation law from the conservation laws dataset will exhibit no shocks, shocks, or rarefactions. Specifically, on this dataset the numerical output achieves a test error of 1.41\% (see Table \ref{test_numeric}). In Figure \ref{fig:test_sol}, the last three rows illustrate that, when present, the locations and intensities of rarefactions and shocks are correctly identified. Finally, Table \ref{tab:description_comparison} shows that the text descriptions accurately capture shocks and rarefactions when present, with an F1 score greater than 0.94.



\subsection{Main Contributions.} We proposes a novel multimodal framework that integrates numerical and textual input and output data for PDE foundation modeling. Our key contributions are as follows:

\begin{itemize}
    \item \textbf{Multimodal Framework:} Our model integrates numerical inputs (e.g., initial conditions and parameters) with text descriptions of physical processes, allowing us to encode contextual information about the underlying dynamics. Moreover, by combining multimodal inputs with multimodal outputs (numerical solutions and text descriptions), our framework provides a novel, comprehensive, and interpretable modeling approach. 

    \item \textbf{Custom Tokenization and Encoding:} Recognizing the potential limitations of GPT models in numerical tasks, we enhance the approach with a custom tokenizer to encode multimodal inputs into a unified token sequence. Textual data are tokenized using a pretrained GPT-2 backbone, while numerical data is encoded through a small multilayer perceptron (MLP). Continuous numerical encodings have also been proposed in \cite{golkar2023xval}, but without the use of an MLP. This encoding approach ensures compatibility between modalities and supports operator learning tasks sensitive to numerical accuracy.

    \item \textbf{Transformer-Based Operator Learning:} We employ a cross-attention-based transformer for numerical output generation, allowing the model to approximate solution operators efficiently. The numerical output decoder is designed to evaluate solutions at independent query points, ensuring scalability with respect to both time and space complexity.

    \item \textbf{Scientific Description Generation:} The model generates textual outputs using the GPT-2 backbone in an autoregressive manner. This enables the generation of descriptive and interpretable explanations of the system's behavior, including properties of the equation, of the solution, possible alternative methods to approximate the solution, etc.

\end{itemize}

\section{Related Works}

\subsection{Foundation Models}
Foundation models are large-scale, pre-trained models that can be fine-tuned for various downstream tasks in natural language processing~\cite{brown2020language,touvron2023llama}, computer vision~\cite{ramesh2021zero}, and other domains including robotics and biology \cite{firoozi2023foundation, zhang2024scientific}. Despite their versatility, these models are not inherently suitable for number-sensitive tasks, such as scientific computing \cite{wang2024recent}, PDE discovery~\cite{schaeffer2017learning}, and time series forecasting \cite{tan2024language}, where high-precision solutions are critical.
Early attempts have been made to adapt foundation models or similar transformer-based structures for scientific computing tasks. Some approaches utilize pre-training and fine-tuning techniques~\cite{chen2024data,herde2024poseidon}, though these methods often require additional computational cost for downstream tasks.
The In-Context Operator Network (ICON) \cite{yang2023context, yang2023fine, yang2024pde,cao2024vicon} uses in-context learning to learn operators through example input-output pairs, which are few-shot learners and have demonstrated generalization capabilities for various differential equations \cite{yang2024pde}. Zero-shot PDE foundation models integrate additional information to aid the prediction process. ICON-LM \cite{yang2023fine} enables in-context learning with both text descriptions of governing physics and numerical data in the model inputs, but outputs are numerical only. In this way, text is used as a label to signal which PDE or task is needed. The PROSE framework~\cite{liu2024prose,sun2024towards,liu2024prosefd,jollie2024time} is a multimodal approach that encodes numerical data along with symbolic representations of the PDE. In recent work, PROSE has been shown capable of extrapolation \cite{sun2024towards} and improved generalization through fine-tuning \cite{sun2024lemon}.
Other methods achieve zero- or few-shot generalization by embedding PDE structures within the network architecture~\cite{lorsung2024physics,ye2024pdeformer}.

\subsection{Multimodal Machine Learning}
Multimodal machine learning focuses on models capable of integrating data from various modalities \cite{lu2019vilbert, sun2019videobert, tan2019lxmert,xu2023multimodal}. A key area of interest in this field is developing methods for information fusion across modalities, analyzing their interactions, and designing appropriate models and algorithms. For instance, in visual-language reasoning \cite{tan2019lxmert, sun2019videobert, li2019visualbert}, combining visual data like images or videos and linguistic information such as captions or text descriptions \cite{tan2019lxmert} facilitates the development of models with enhanced multimodal understanding \cite{li2019visualbert}. Similarly, AI robots utilize multimodal sensors, including cameras, radar, and ultrasound, to interpret their environments and make well-informed decisions \cite{feng2020deep, liu2021multimodal}. Notably, the concept of a multimodal sentence was introduced in PaLM-E \cite{driess2023palm}, where image and text can appear anywhere in the sentence and be processed in a flexible manner.

\subsection{Transfer Learning}
Transfer learning has become a cornerstone of modern machine learning, enabling the adaptation of pre-trained models to specialized tasks with relatively limited data \cite{pan2009survey}. In the context of large language models (LLMs), such as GPT and BERT, fine-tuning on domain-specific data allows the model to leverage its extensive pre-trained knowledge while focusing on the nuances of a target domain \cite{radford2019language, kenton2019bert}. For instance, fine-tuning GPT-2 or GPT-3 models on scientific texts has shown improved performance in tasks such as equation generation, technical summarization, and scientific question answering \cite{brown2020language}. In our work, which follows the operator learning paradigm, we fine-tune an LLM with examples of numerical equations, symbolic representations, and descriptive text to enhance its ability to generate meaningful and contextually relevant outputs. The inclusion of this step in our framework, aligns well with recent advancements in multimodal tasks \cite{han2021pre,liu2024prose,sun2024towards}, demonstrating the importance of adapting pre-trained LLMs to specialized contexts.

\section{Methodology}

\begin{figure}[t]
    \centering
    \includegraphics[width=\linewidth]{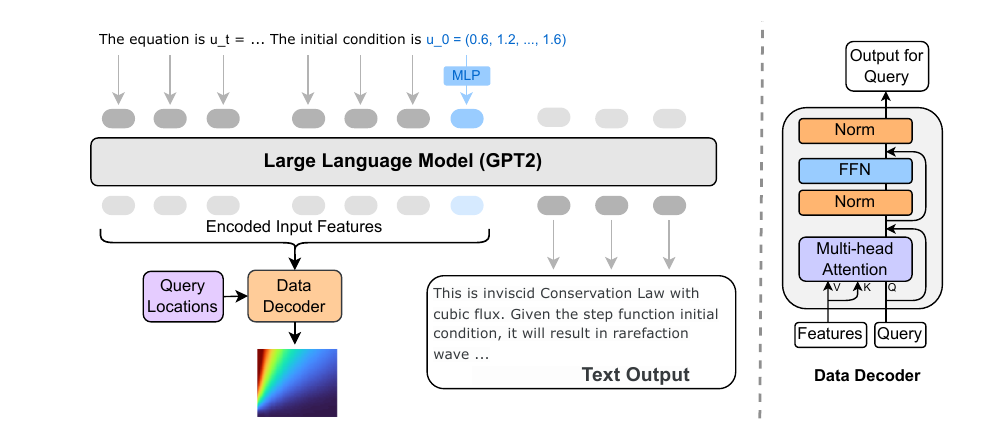}
    \caption{Model Illustration.  Our model processes multimodal input, where textual prompts describe equations, and numerical values represent initial conditions and parameters. A custom tokenizer encodes text using a GPT-2 tokenizer and numerical inputs via an MLP. The tokenized sequence is processed by an LLM backbone, followed by two decoding pathways: a transformer decoder for text generation and a data decoder with cross-attention to construct the operator.}
    \label{fig:model}
\end{figure}

\subsection{Model Overview}
The model consists of two main components: (1) pretrained GPT-2 backbone \cite{radford2019language}, and (2) cross-attention-based transformer for operator-style output evaluation similar to the one used in \cite{liu2024prose}. An illustration of the model is provided in Figure \ref{fig:model}. Given multimodal input data containing textual prompts describing the equation and numerical values representing initial conditions and parameters, our custom tokenizer first encodes the input data into a multimodal sequence of mixed tokens. The text input is encoded similarly to language models, where the GPT-2 tokenizer is used. For the numerical inputs, a small multilayer perceptron (MLP) is used to encode the numerical data into feature vectors. This continuous encoding strategy similar to the one proposed in \cite{golkar2023xval}, demonstrates improved generalization, and is more suitable for number-sensitive tasks. 
The tokenized sequence is fed into the LLM backbone, which processes the input sequence and performs token mixing. Two different decoding methods are used to generate bi-modality output, and we discuss them in the following sections. 

\subsection{Transformers}
Transformers are attention-based models that excel in tasks involving sequential data \cite{vaswani2017attention}. The attention mechanism \cite{bahdanau2014neural} is the key component of transformers, which is used to weigh the importance of different elements in a sequence, allowing efficient parallelization and capture of long-range dependencies. This architecture has been highly successful in various domains, including natural language processing \cite{radford2019language}, computer vision \cite{dosovitskiy2020image}, and more recently scientific computing \cite{cao2021choose,yang2023context,liu2024prose}. It is also the building block of the GPT-2 backbone used in our model. 
Each layer in standard decoder-only transformers such as GPT-2 consists of two sub-layers: self-attention and feedforward layers. Given the input sequence $X = (x_1,x_2,\dots,x_n)\in \mathbb{R}^{n\times d}$, each transformer layer processes the input as follows:
\begin{align}
    X' &= \text{LayerNorm}(X)\\
    Y &= X + \text{MultiheadAttention}(X',X',X')\\
    \text{Output} &= Y + \text{FeedForward}(\text{LayerNorm}(Y))\in \mathbb{R}^{n\times d}.
\end{align}
The feedforward layer is a 2-layer multilayer perceptron (MLP), and the multihead attention mechanism (with $h$ heads) is defined as follows:
\begin{align}
    \text{Attention}(Q, K, V) &= \text{softmax}\left(\frac{QK^T}{\sqrt{d_K}}\right)V, \\
    \text{MultiheadAttention}(Q, K, V) &= \text{Concat}(\text{head}_1, \ldots, \text{head}_h)W^O,
\end{align}
where $d_K$ is the number of columns of $K$, softmax is computed along each row, and each head is computed via
\begin{align}
    \text{head}_i &= \text{Attention}(QW_i^Q, KW_i^K, VW_i^V)\in \mathbb{R}^{n\times (d/h)}.
\end{align}
Here, \(Q\), \(K\), and \(V\) the input are the query, key, and value matrices of dimension $n\times d$, respectively (for self-attention, $Q=K=V=X$). \(W_i^Q,W_i^K,W_i^V\in \mathbb{R}^{d\times(d/h)}\) are the learned projection matrices for the \(i\)-th head, and \(W^O\in \mathbb{R}^{d\times d}\) is the output projection matrix.

\subsection{Operator Evaluation}
We generate numerical outputs representing equation solutions in the operator style. The Data Decoder in Figure \ref{fig:model} employs a cross-attention mechanism to construct the operator, creating a connection between the LLM-processed input sequence and the output functions. The query locations, which represent the independent variables of these output functions, act as evaluation points. Notably, these query locations are independent of each other, meaning that evaluating the operator at one point does not influence its evaluation at another point. Consequently, the time and space complexity scale linearly with the number of query locations. Furthermore, since the evaluation points are decoupled from the network's encoding process, this approach resembles the principles underlying branch and trunk networks in DeepONet \cite{lu2019deeponet}.

\subsection{Autoregressive Text Generation}
To generate text output, we use the standard autoregressive next-token prediction approach with the LLM backbone \cite{vaswani2017attention}. To generate the complete text output, we iteratively predict the next token in the sequence until a special end-of-sequence token is generated. At each step, the LLM generates the probability distribution for the next token, and we greedily select the token with the highest probability. Note that other decoding strategies such as beam search or sampling can be used to improve the quality of the generated text \cite{kulikov2018importance, ott2018analyzing}. During training time, to enable parallelization, we input the complete text sequence and explicitly mask out the future tokens for text generation and output tokens for Data Decoder. 

\subsection{Loss Function}

For the operator evaluation, we use the relative squared error $\mathcal{L}_n$, which makes learning more uniform among different families of equations with various data scales as shown in \cite{jadon2024comprehensive}. We also pad the equation solution to unify the data dimension so that the padded positions are not included in the loss calculation. For text generation, we use the standard cross-entropy loss $\mathcal{L}_t$ between the generated text and the ground truth text. The total loss is the sum of the two losses:
\begin{align*}
    \mathcal{L} &= \mathcal{L}((\bm{u}, \bm{s}), (\hat{\bm{u}}, \hat{\bm{s}}))= \alpha \mathcal{L}_n(\bm{u}, \hat{\bm{u}}) + \beta \mathcal{L}_t(\bm{s}, \hat{\bm{s}})\\
    &= \frac{\alpha}{B}\sum_{i=1}^B \frac{\norm{\bm{u}_i - \hat{\bm{u}}_i}_2^2}{\norm{\bm{u}_i}_2^2} + \frac{\beta}{B}\sum_{i=1}^B \text{CrossEntropy}(\bm{s}_i, \hat{\bm{s}}_i),
\end{align*}
where the weights $\alpha,\beta$ are hyperparameters to balance the two losses, $i$ is the index for $i$-th element in the batch of size $B$, $\bm{u}$ is the ground truth numerical output, $\hat{\bm{u}}$ is the model numerical output, $\bm{s}$ is the text output ground truth, and $\hat{\bm{s}}$ is the model text output logits.

\section{Results and Discussion}\label{sec:results}
This section presents the results of numerical experiments. In the first section, we describe the evaluation metrics, then we conduct four studies. First, we show the results on test data sampled from the same distribution as the training data. Second, we study the text generation performance for each equation class. Then we perform a study on out-of-distribution data. Finally,  we test the ability of our network to extrapolate dynamics in time. 

In all numerical experiments (Sections \ref{sec:test_numeric},\ref{sec:OOD}, \ref{sec:Extrap}), the model receives a multimodal input, combining textual and numerical initial conditions and parameters, and is tasked with producing the numerical solution over the full time interval. While the model generates both numerical solutions and text descriptions, these sections only evaluate the model's numerical prediction capabilities. The text description output is evaluated separately in Section \ref{sec:text_gen}. In the first two numerical studies (in-distribution and out-of-distribution test data) we provide the initial condition (solution at time 0) and task the model to predict the solution on the time interval $[0,5]$. For the extrapolation in time study,the model is given its prediction of the solution at time 5 as the initial condition and asked to predict the solution on the longer time interval $[5,10]$. 

The results in this section demonstrate that our model not only provides accurate solutions for equations where parameters are sampled within the same intervals as the training data, but also that it maintains a reasonable predictive accuracy for out-of-distribution (OOD) cases where parameters are sampled from larger intervals and, in some cases, is capable of extrapolating solutions in time. Furthermore, Section \ref{sec:text_gen} demonstrates that our model is also able to produce accurate and consistent text descriptions.

\subsection{Dataset Overview}\label{data}
We generate a synthetic dataset consisting of 52 parametric differential equations, including both ODEs and PDEs of varying dimensions. The dataset is designed to cover a wide range of dynamics, including linear and non-linear systems, conservation laws, and reaction-diffusion equations. For each of the 52 parametric equations, we sample 100 parameters in the range $[Q-0.1Q, Q+0.1Q]$, where $Q$ denotes the value of interest, resulting in a total of 5200 distinct equations. In addition to solution trajectories, each family of equations also comes with a set of text descriptions that describe the dynamics and properties of the system, generated using GPT-4 \cite{achiam2023gpt}. We enumerate all 52 parametric equations in Table~\ref{equations}. For more details about the dataset, we refer to Appendix \ref{sec:dataset_details}.

\begin{table}[ht]
\centering
\begin{tabular}{|c|l|c|}
\hline
\textbf{Type} & \textbf{Equation} & \textbf{Index} \\
\hline
\multirow{5}{*}{1D ODE} 
& $ u_t = a \sin(2 \pi t)  u $ & 1 \\
& $ u_t = a \exp(-t) + b $ & 2 \\
& $ u_t = a  t^2 \cos(u) + b  u $ & 3 \\
& $ u_t = a \sin(\exp(-0.5 t) \sin(3 \cdot t)) + b  u $ & 4 \\
& $ u_t = a  t \sin(u) $ & 5 \\
\hline
\multirow{5}{*}{2D ODE} 
& Van der Pol & 8 \\
& Lotka-Volterra & 9 \\
& FitzHugh-Nagumo & 10 \\
& Brusselator & 11 \\
& Duffing & 12 \\
\hline
\multirow{2}{*}{3D ODE} 
& SIR Model& 6 \\
& Neural Dynamics& 7 \\
\hline
\multirow{12}{*}{PDE} 
& Heat & 13 \\
& Porous Medium & 14 \\
& Klein Gordon & 15 \\
& Sine Gordon & 16 \\
& Cahn Hilliard & 17 \\
& Korteweg De Vries & 18 \\
& Advection & 19 \\
& Wave & 20 \\
& Diffusion-reaction Logistic & 21 \\
& Diffusion-reaction Linear & 22 \\
& Diffusion-reaction Bistable & 23 \\
& Diffusion-reaction Square Logistic & 24 \\
& Fokker-Plank & 34 \\
\hline
\multirow{18}{*}{Conservation Laws} 
& Burgers & 25 \\
& Inviscid Burgers & 26 \\
& Conservation law Linear Flux & 27 \\
& Conservation law Cubic Flux & 28 \\
& Inviscid Conservation law Cubic Flux & 29 \\
& Conservation law Sine Flux & 30 \\
& Inviscid Conservation law Sine Flux & 31 \\
& Conservation law Cosine Flux & 32 \\
& Inviscid Conservation law Cosine Flux & 33 \\
& Burgers-Inviscid Conservation law Cosine Flux with one shock & 35-43 \\
& Burgers-Inviscid Conservation law Cosine Flux with rarefaction & 44-52 \\
\hline
\end{tabular}
\caption{Equations by type with corresponding indices. Each equation has parameters (such as $a$ and $b$ in the 1D ODE set) that are sampled during the data generation process.}
\label{equations}
\end{table}

\subsection{Evaluation Metrics}
We evaluate performance using a {relative error} for numerical outputs and the {BERTScore} for text outputs. These metrics are defined as follows:
\begin{enumerate}
    \item Given a numerical prediction \(\hat{\bm{u}}\) and ground truth solution \(\bm{u}\), the relative error is defined as:  
    \[
    \text{Relative Error} = \frac{\|\hat{\bm{u}} - \bm{u}\|}{\|\bm{u}\|}
    \]
    where \(\|\cdot\|\) denotes the Euclidean norm.  

    This metric measures the accuracy of the numerical solution generated by the model over the time interval compared to the true solution. The structure of \(\bm{u}\) depends on the problem dimension:  
    \begin{itemize}  
        \item For 1D ODEs, \(\bm{u}\) is a vector containing solution values at discrete query time points.  
        \item For higher-dimensional ODEs, \(\bm{u}\) is a matrix where each row corresponds to a different component of the solution, and each column represents a query time point.  
        \item For 1D PDEs, the solution \(\bm{u}\) has spatial structure. We discretize the spatial domain into 128 points and treat each spatial point as an independent dimension. Thus, the solution is represented as a matrix of size \(128 \times\) (number of query time points), where each row corresponds to a specific spatial location, and each column represents a different query time.  
    \end{itemize}  

\item {BERTScore} evaluates sentence similarity by comparing token-level representations using a pre-trained BERT model. Precision, recall, and F1 scores are calculated based on token alignments between two sentences. {Precision} measures the proportion of tokens in the generated sentence \(\hat{x}\) that are relevant to the reference sentence \(x\), indicating how much of the generated content matches the reference. {Recall} measures the proportion of tokens in the reference sentence \(x\) that are found in the generated sentence \(\hat{x}\), reflecting how well the generated sentence captures the content of the reference. The {F1 score} is the harmonic mean of precision and recall, balancing relevance and coverage. An F1 score of 1 indicates perfect precision and recall, while lower scores suggest imbalances between the two. Precision, recall, and F1-score are computed as follows:

\[
R_{\text{BERT}} = \frac{1}{|x|} \sum_{x_i \in x} \max_{\hat{x}_j \in \hat{x}} x_i^\top \hat{x}_j
\]
\[
P_{\text{BERT}} = \frac{1}{|\hat{x}|} \sum_{\hat{x}_j \in \hat{x}} \max_{x_i \in x} x_i^\top \hat{x}_j
\]
\[
F_{\text{BERT}} = \frac{2 \cdot P_{\text{BERT}} \cdot R_{\text{BERT}}}{P_{\text{BERT}} + R_{\text{BERT}}}
\]
where \(x_i\) and \(\hat{x}_j\) are the token embeddings from the reference \(x\) and candidate \(\hat{x}\), respectively, and \(x_i^\top \hat{x}_j\) represents their dot product.
\end{enumerate}

\subsection{Numerical Predictions on Test Data}\label{sec:test_numeric}
In this section, we present numerical results on test data sampled from the same distribution as the training data. As detailed in Section \ref{data}, to generate the training data, we uniformly sample the parameters of each differential equation within the range \([Q-0.1Q, Q+0.1Q]\), where \(Q\) denotes the parameter value of interest. This range corresponds to a \(10\%\) relative variation around each parameter's nominal value. In this section, we evaluate the in-distribution performance of our method, i.e., its performance on test data where the parameters are also uniformly sampled from the range \([Q-0.1Q, Q+0.1Q]\). 

\begin{figure}[h!]
    \centering
    \includegraphics[width=0.96\linewidth]{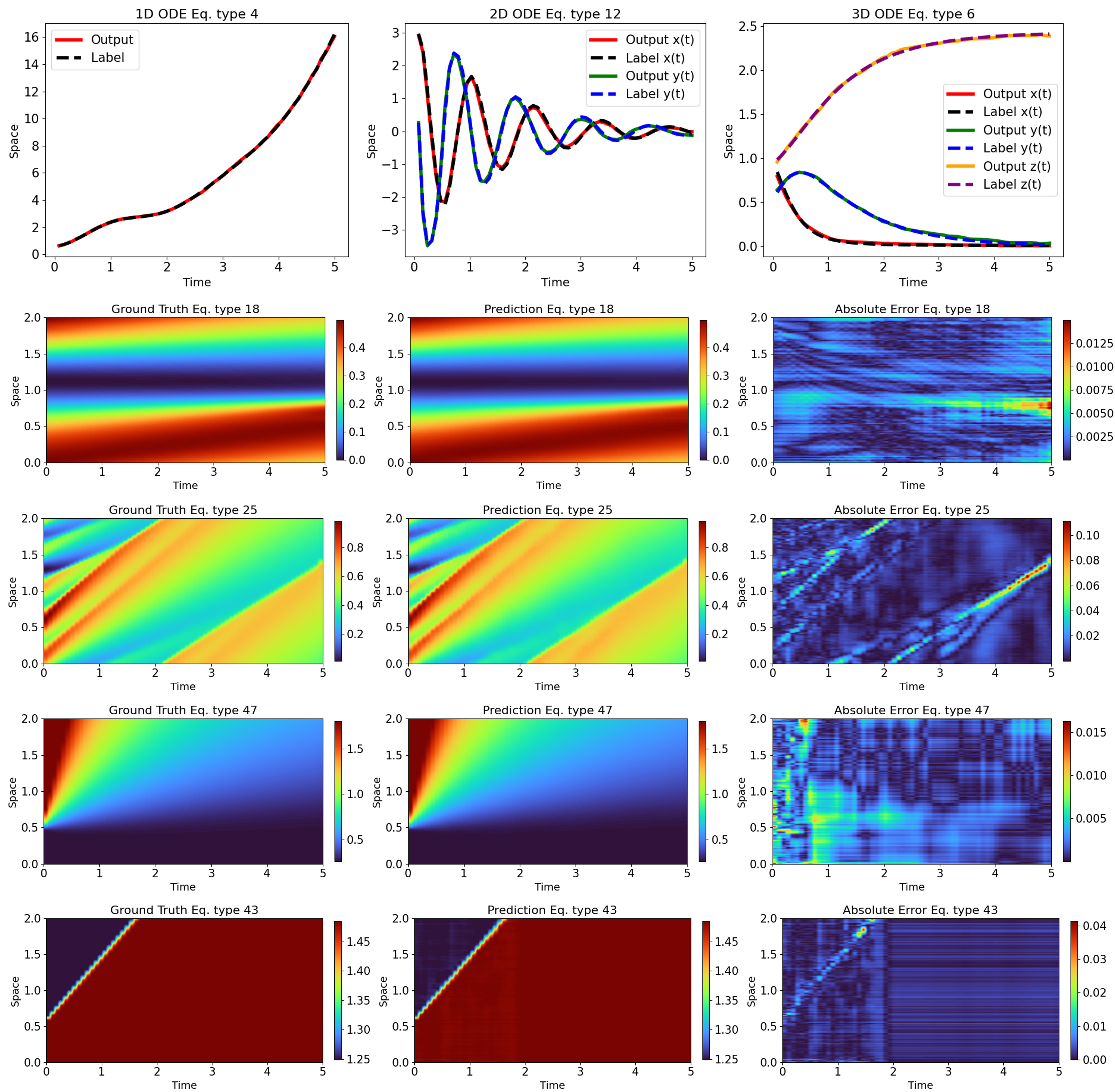}
    \caption{Comparison of Outputs: For PDE examples, we show from left to right the ground truth solution, the predicted solution, and the absolute difference. \textbf{First row:} from left to right 1D ODE (index 4), 2D ODE (index 12, Duffing system), 3D ODE (index 7, Neural Dynamics). \textbf{Second row:} PDE (index 18, Korteweg De Vries equation). \textbf{Third row:} Conservation Law (index 25, Burgers' equation). \textbf{Fourth row:} Conservation Law with rarefaction (index 47, Inviscid Conservation law Cubic Flux); \textbf{Fifth row:} Conservation Law with shock (index 43, Inviscid Conservation law Cosine Flux)}
    \label{fig:test_sol}
\end{figure}

Table \ref{test_numeric} shows the results for each equation class. We can see that in all cases the low relative error is achieved ($<5.4\%$) with errors especially low for the PDEs and Conservation Laws classes ($<1.9\%$). This discrepancy between ODE and PDE sets can be explained by the fact that low-dimensional systems often have simpler dynamics, making them more sensitive to perturbations and modeling inaccuracies. This sensitivity can make small deviations more pronounced and result in larger relative errors.
Figure \ref{fig:test_sol} shows one prediction example per class. As explained above, for testing the only numerical data provided to the model was the initial condition and the equation parameters in the multimodal input sentence and the model was tasked to predict the solution on $[0,5]$. From the figure, we see that the prediction is almost indistinguishable from the ground truth. For the PDE and Conservation laws examples the largest error is observed where there is a sharp transition in the dynamics. However, in all cases, the main features of the solutions are correctly predicted. Notably, in the case of conservation laws with rarefaction and shocks (rows 4 and 5 of the figure), the location of both the rarefaction and shock are correctly identified. 

\begin{table}[t]
\centering
\begin{tabular}{|c|c|}
\hline
\textbf{Class} & \textbf{Relative Test Error (\%)} \\
\hline
1D ODE & 3.00\% \\
2D ODE & 5.36\% \\
3D ODE & 4.43\% \\
PDE & 1.85\% \\
Conservation Laws & 1.41\% \\
\hline
\textbf{Total Average} & 3.21\% \\
\hline
\end{tabular}
\caption{Relative Errors on Test Data per Class}
\label{test_numeric}
\end{table}

\subsection{Scientific Text Description Generation}\label{sec:text_gen}
Each equation in our dataset was associated with 50 different text descriptions. During training, given the equation in text form and the initial condition as numerical data, the model produces a text description. In this work, the text descriptions for training were generated using GPT-4 and described multiple properties of the equation to be solved.  Some descriptions focus on the equation and its properties (such as order, linearity, and main equation terms), some focus on classical numerical methods which may be used as an alternative to produce an approximate solution, and some focus on the natural dynamics that the equation describe (for example the Lotka Volterra system describes the interactions between prey and predators in an ecosystem). For evaluation, we accept as ``correct'' any description that is consistent with the corresponding equation.

To evaluate the accuracy of generated text descriptions, we use BERTScore to measure sentence similarity. Specifically, for a given input, we generate an output text description and compare it with a randomly selected text description from the set of 50 associated with the equation. The results of this comparison on test data for each class are shown in Table \ref{BERTscore}. From the table, we see that the average BERTscore is high ($>0.935$) for all classes. Note that the BERTScore is slightly below a perfect match because of the inherent variability in the text descriptions. When we randomly select a description for comparison, it may focus on a different property of the equation than the one described by the generated output. For example, one description may emphasize the order of the equation, while another focuses on a classical numerical method used to approximate its solution. Although both descriptions are consistent and correct for the given equation, their semantic similarity would be low due to the differences in focus. This discrepancy leads to slightly reduced BERTScores, even when the generated descriptions are valid. For 100 test examples, we manually verified that the text descriptions were consistent with the provided multimodal input and found them to be accurate 100\% of the time.

Table \ref{tab:description_comparison} shows the input data, example description for that class, generated text, and corresponding BERTscore-F1 for the numerical results from Section \ref{sec:test_numeric}, Figure \ref{test_numeric}. We see that in all cases the generated description is consistent with the input data, i.e., it correctly describes the equation or the properties of the solution. However, the BERTScore-F1 is not perfect because of small semantic discrepancies between the example and generated descriptions. For example in the 1D ODE case the example description mentions \textit{``a decaying exponential term"} while the generated text mentions \textit{``a time-varying control term"}. Both of these descriptions are consistent and correct with respect to the input data (\textit{``The ODE is $u_t = a\sin(\exp(-0.5t)\sin(3t)) + bu$ we have initial data $u =...$ and coefficients $[a,b] = ...$"}), but semantically they are different, hence the reduced BERTScore of 0.911. 

An interesting scientific application of our method can be seen when we apply it to conservation laws (see the last 3 rows of Table \ref{tab:description_comparison}). In that case, we are interested in generating text descriptions that can characterize the solution behavior and determine whether shocks or rarefactions are likely to form based solely on the input data.  This capability of our method is particularly significant as being able to predict whether shocks or rarefactions will form directly from input data is crucial for understanding and controlling systems governed by conservation laws, especially when dealing with complex, non-linear dynamics arising for example in fluid dynamics, traffic flow, and material science applications. This example shows that our multimodal LLM method is able to leverage physical knowledge encoded in equation structures (text input data) alongside numerical data (such as initial conditions and parameters) to produce informed conclusions (determining the presence of shocks or rarefactions).
Finally, the ability of our model to produce as output simultaneously a numerical solution (see Figure \ref{test_numeric}) and a text description (see Table \ref{tab:description_comparison}) allows for a deeper understanding of the problem and more interpretable results.

These examples demonstrated the ability of our model to produce any description coherent with the input data without requiring the description to focus on a specific property of the equation or of the solution. A simple modification of this model could be to include in the input data a short text prompt that specifies what property of the problem the output text description should focus on. For example, we could require a description of the long-time behavior of the solution, a possible discretization of the equation, or if the given equation can be used to model real-world dynamics. We leave this as future work as this would require a more general example description set.

\begin{table}[t]
\centering
\begin{tabular}{|c|c|c|c|}
\hline
\textbf{Class} & \textbf{BERTScore - Precision} & \textbf{BERTScore - Recall} & \textbf{BERTScore - F1} \\ \hline
1D ODE & 0.919 & 0.915 & 0.917 \\ \hline
2D ODE & 0.950 & 0.950 & 0.950 \\ \hline
3D ODE & 0.918 & 0.915 & 0.916 \\ \hline
PDE & 0.947 & 0.949 & 0.948 \\ \hline
Conservation Laws & 0.952 & 0.952 & 0.952 \\ \hline
\textbf{Total Average} & 0.937 & 0.936 & 0.937 \\ \hline
\end{tabular}
\caption{BERTScore results per class}
\label{BERTscore}
\end{table}

\subsection{Out-of-distribution Testing Performance}\label{sec:OOD}
In the previous section, the train and test datasets were generated with parameters uniformly sampled from the range \([Q-0.1Q, Q+0.1Q]\) where $Q$ was the quantity of interest. In this section, we test our algorithm on out-of-distribution data, specifically, the test data parameters are sampled from \([Q-0.2Q, Q+0.2Q]\) (20\% relative range) and from \([Q-0.3Q, Q+0.3Q]\) (30\% relative range).  Table \ref{OOD_test} shows the test errors respectively for 10\% relative range (in-distribution test data), 20\%, and 30\% relative range (out-of-distribution test data). As expected, we observe an increase in the relative error as we increase the parameter's relative range; however, the error remains low for most equation classes. The average error is $<3.3\%$ for the 10\% range, $<7.8\%$ for the 20\% range, and $<12.0\%$ for the 30\% range.

We note that the 2D ODE class exhibits the largest increase in error as we expand the parameter range. This behavior is due to the sensitivity of these systems to changes in their parameters. Many of these equations, such as the Van der Pol oscillator, Lotka-Volterra model, FitzHugh-Nagumo model, Brusselator, and Duffing oscillator, exhibit highly nonlinear and often oscillatory dynamics. In such systems, even small deviations in parameter values can result in significant changes in the amplitude, frequency, or overall trajectory of the solutions leading to dynamics that deviate substantially from those seen during training. For example, for the Duffing oscillator small parameter shifts can move the system between periodic and chaotic behaviors while for Van der Pol small changes in parameters can shift the system between weakly and strongly oscillatory regimes. 

As the relative parameter range increases, we observe that the error roughly doubles. Since our model relies solely on the initial condition as numerical input, it must infer the entire trajectory based on limited information, which can amplify errors when the system's behavior deviates significantly from the training data. A possible way to mitigate this issue is to provide the model with additional time steps of the solution as input. Including more time steps offers the model a richer context about the dynamics, allowing it to identify trends and patterns that remain consistent even as parameters vary. This additional information can stabilize and regularize the learning process, reducing the model's reliance on extrapolation from the initial condition and improving its ability to adapt to unseen parameter values \cite{liu2024prose}. We leave the addition of multiple time steps as input as a future work. 

\begin{table}[H]
\centering
\begin{tabular}{|c|c|c|c|}
\hline
\textbf{Class} & \textbf{10\% Relative Range (Training)} & \textbf{20\% Relative Range} & \textbf{30\% Relative Range} \\
\hline
1D ODE & 3.00\% & 7.19\% & 12.56\% \\
2D ODE & 5.36\% & 15.15\% & 25.71\% \\
3D ODE & 4.43\% & 6.46\% & 8.86\% \\
PDE & 1.85\% & 3.56\% & 5.82\% \\
Conservation Laws & 1.41\% & 3.54\% & 6.66\% \\
\hline
\textbf{Total Average} & 3.21\% & 7.78\% & 11.92\% \\
\hline
\end{tabular}
\caption{Relative Errors on out-of-distribution test data. From left to right, results on 10\% relative range (in-distribution data), and results from 20\% and 30\% relative range for each equation class.}
\label{OOD_test}
\end{table}

\begin{table}[H]
\centering
\renewcommand{\arraystretch}{1.5} 
\begin{tabular}{|m{1.9cm}|m{5.3cm}|m{3.8cm}|m{3.8cm}|m{1cm}|}
\hline
\textbf{Class\newline(Index)} & \textbf{Input Data} & \textbf{Example Description} & \textbf{Generated Text} & \textbf{F1\newline Score} \\ \hline

1D ODE\newline(Index 4) & 
The ODE is $$u_t = a\sin(\exp(-0.5t)\sin(3t)) + bu$$ we have initial data $u =...$ and coefficients $[a,b] = ...$
 &
This first-order linear equation includes a sine function affected by a decaying exponential term and is linear in \( u \). &
The ODE is characterized by a first-order linear term with a sine function influenced by a time-varying control term. &
0.911 \\ \hline

2D ODE\newline(Index 12) &
The Duffing system is 
\begin{align*}
    x_t = y,\\
    y_t = -\delta y - \alpha x - \beta x^3.
\end{align*}
The initial data is $[x,y] = ...$, and the parameters are $[\alpha, \beta, \delta] = ...$ &
The Duffing system, with \(\alpha\), \(\beta\), and \(\delta\), shows how cubic stiffness and damping lead to non-linear oscillations in \( x \) and \( y \). &
In the Duffing system, parameters \(\alpha\), \(\beta\), and \(\delta\) illustrate how cubic stiffness and damping lead to non-linear oscillatory behavior in \( x \) and \( y \). &
0.977 \\ \hline

3D ODE\newline(Index 7) &
The Neural Dynamics system is 
\begin{align*}
    E_t = \alpha E - \beta  E  I - \gamma  E + 0.01 \sin(t),\\
    I_t = \delta  E - \epsilon  I,\\
    H_t = \theta I - \phi H.
\end{align*}
We have initial data $[E,I,T]=...$, parameters $[\alpha, \beta, \gamma, \delta] = ...$
 &
This model describes how sinusoidal input influences neural excitatory and inhibitory balance. &
This model describes how excitatory and inhibitory neural signals evolve under sinusoidal stimulation. &
0.960 \\ \hline

PDE\newline(Index 18) &
The equation is $$u_t + q^2\,u_{xxx} + u\,u_x = 0$$ with observed initial data $u= ...$ and coefficient $q = ...$
 &
The equation of Korteweg–de Vries captures wave propagation in shallow channels, influenced by nonlinearity and dispersion. &
The equation of Korteweg–de Vries describes the behavior of solitons in shallow water, balancing nonlinear and dispersive forces. &
0.945 \\ \hline

Conservation Law\newline(Index 25) &
The equation is 
$$u_t = -q_1(f(u))_x + \frac{q_2}{\pi} u_{xx}$$  $f(u) = 1/2\; u^2$. We have initial condition $= ...$ and coefficients $[q_1, q_2] = ...$
 &
The viscous Burgers' equation, with sinusoidal initial conditions, models fluid dynamics that stay smooth without shocks due to viscosity. &
The viscous Burgers' equation models fluid dynamics initiated by sinusoidal waves, ensuring no shocks form due to viscosity. &
0.942 \\ \hline

Conservation Law Rarefaction\newline(Index 47) &
The equation is 
$$u_t = -q_1(f(u))_x + \frac{q_2}{\pi} u_{xx}$$ where $f(u) = 1/3\; u^3$. Initial condition $= ...$ and coefficients $[q_1, q_2] = ...$
 &
The solution of the inviscid conservation law with cubic flux, starting from step function initial conditions, demonstrates a rarefaction wave forming. &
With step function initial conditions, the inviscid conservation law with cubic flux models the fluid flow where a rarefaction wave forms. &
0.941 \\ \hline

Conservation Law Shock\newline(Index 43) &
The equation is 
$$u_t = -k(f(u))_x\,u_{xx}$$
where $f(u) = \cos(u)$. Initial condition  $= ...$ and coefficient $k = ...$
 &
The inviscid conservation law with cosine flux, initialized from a step function, leads to fluid dynamics featuring one shock. &
The inviscid conservation law with cosine flux, starting with a step function, ensures fluid dynamics include one shock. &
0.963 \\ \hline
\end{tabular}
\caption{Example and generated text descriptions and BERTScore-F1 similarity with corresponding input data for different equations.}
\label{tab:description_comparison}
\end{table}

\subsection{Extrapolation in time}\label{sec:Extrap}
In this section, we study the ability of our method to extrapolate the dynamics in time. As explained in Section \ref{data}, the model is trained on data generated on the time interval $[0,5]$. In the previous sections, the trained model received as numerical input the solution at time \(t=0\) (the initial condition) and was tasked with producing the solution over the entire time interval \([0,5]\). In contrast, in this section, we investigate the model's capacity for temporal extrapolation by providing, as the initial condition, its own approximation of the solution at time \(t=5\). The model is then tasked with generating the solution on the time interval \([5,10]\).
Note that we only show extrapolation results for certain classes of equations where the solution at time \(t=5\) is similar to the initial condition used for training. This is because if the solution at \(t=5\) differs significantly from the training data the new input may be an OOD example.

\begin{figure}[h!]
    \centering
    \includegraphics[width=0.99\linewidth]{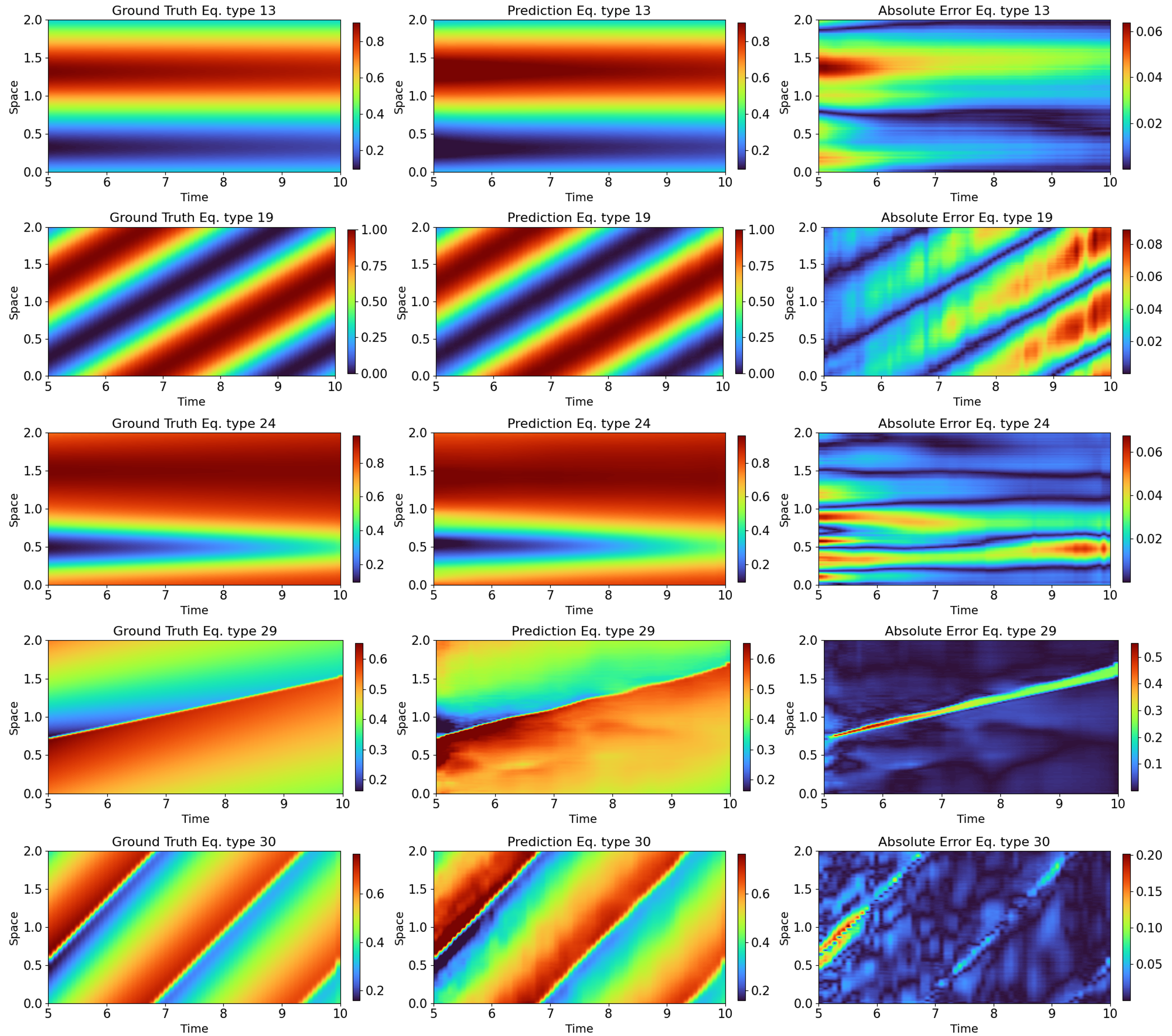}
    \caption{Example of extrapolation in time. The model is trained on data generated for $t\in[0,5]$ and tasked to predict the solution for $t\in [5,10]$. From left to right, ground truth solution, predicted solution, absolute difference. \textbf{First row:} Heat Equation (index 13), \textbf{Second row:} Advection Equation (index 19). \textbf{Third row:} Diffusion-reaction Square Logistic (index 24). \textbf{Fourth row:} Inviscid Conservation law Cubic Flux (index 29) ; \textbf{Fifth row:} Conservation law Sine Flux (index 30)}
    \label{fig:Extrap_sol}
\end{figure}

Table \ref{tab:extrap} shows extrapolation results for heat equation, advection equation, diffusion-reaction equation,
inviscid conservation law with cubic flux and
conservation law with sine flux. For each class, we sample 70 different parameters and initial conditions and report the average relative error in the numerical solutions over the time interval \([5,10]\). We see that the extrapolation error is $< 13\%$ for most equation classes showing the ability of our model to extrapolate in time in most cases. For the advection equation class, the error is quite high (around 30\%) due to the nature of the solution, which often includes sharp changes. While the general shape of the solution can often be predicted, if the initial approximation at \(t=5\) is slightly shifted or inaccurate, these sharp transitions cause a significant increase in the error on the time interval \([5,10]\) (see Figure \ref{fig:Extrap_sol} row 2 for an example of an advection equation extrapolated solution with sharp intensity changes).

Figure \ref{fig:Extrap_sol} shows one example of extrapolated solution per class on the time interval $[5,10]$. By comparing the ground truth (first column of the figure) with the prediction (second column of the figure) we can see that in all cases the general dynamic is correctly extrapolated. The absolute error (last column in the figure) indicates that the largest errors occurs where the solution exhibits sharp changes in intensity. This is expected, as even a minor deviation in the solution at \(t=5\) can lead to a shift in the prediction, resulting in a large absolute difference. This occurs even when the average error remains small and the main features of the dynamics are accurately captured.

\begin{table}[H]
\centering
\begin{tabular}{|l|c|}
\hline
\textbf{Equation} & \textbf{Relative Error (\%)} \\
\hline
Heat Equation & 8.65\% \\
Advection Equation & 29.3\% \\
Diffusion-Reaction Square Logistic & 9.08\% \\
Inviscid Conservation Law Cubic Flux & 7.94\% \\
Conservation Law Sine Flux & 12.9\% \\
\hline
\end{tabular}
\caption{Relative Errors for extrapolation in time for different equations classes. The model is given as numerical input its approximation of the solution at time $t=5$ and tasked to produce the solution on the time interval $[5,10]$.}
\label{tab:extrap}
\end{table}

\noindent
\textbf{Comparison with Standard Operator Learning}: 
Popular operator learning methods, such as Fourier Neural Operators (FNO) \cite{li2020fourier} and DeepONet \cite{lu2021learning}, assume a fixed underlying equation across training samples and learn mappings from function spaces to function spaces using only numerical inputs. In contrast, our network encodes multiple equations simultaneously, with only initial conditions and equation parameters provided as numerical data (while equations are encoded in the text input). This yields a problem ill-posed for standard operator learning methods, which would have to infer solely from the numerical input both the solution operator and governing equation (including unknown forces), making a fair comparison not feasible. As a baseline test, \cite{liu2024prose} (Table 5) compared FNO and DeepONet with the PROSE model applied to ODE examples, where historical context is provided, i.e., multiple input snapshots of the solution are available instead of only the initial condition. Even with this additional information, DeepONet and FNO produced relative errors respectively of 8.51\% and 9.40\%. Similarly, in \cite{sun2024towards} (Table 5), FNO and DeepONet are compared to the PROSE-PDE model on the same PDE dataset used in this paper (noting that \cite{sun2024towards} uses multiple input snapshots, while we only use the initial condition). For the PDE examples, the relative errors for DeepONet and FNO were respectively 32.07\% and 34.86\%.

\section{Conclusion}
This work introduces a multimodal transformer-based framework for operator learning that integrates numerical and textual data to approximate solution operators for a diverse class of ODEs and PDEs. By incorporating both types of data, our model bridges the gap between numerical predictions and scientific interpretability, making it well-suited for applications where numerical and textual inputs/outputs are required. The framework’s ability to output both numerical solutions at query locations and scientifically meaningful text descriptions of system dynamics represents a step toward interpretable and comprehensive PDE foundation modeling. 

Our experiments demonstrated that the proposed approach is able to produce accurate solutions for in-distribution data, with average relative error less than 3.3\%,  and out-of-distribution data, with average relative error less than 7.8\%, showcasing its robustness and generalization capabilities. Notably, the model can also extrapolate solutions over extended time intervals. Finally, the generated text descriptions are both accurate and informative, with coherent descriptions generated 100\% of times, covering diverse aspects of the equations and their solutions, including physical interpretations and alternative numerical methods to approximate solutions. Future work could explore extending the framework to more complex systems, incorporating additional data modalities, improving scalability for high-dimensional systems, and expanding the scientific text description scope.

\subsection*{Acknowledgments}

Elisa Negrini, Yuxuan Liu, and Hayden Schaeffer are supported in part by AFOSR MURI FA9550-21-1-0084, NSF DMS 2427558, and NSF DMS 2331033.
Liu Yang is supported by the NUS Presidential Young Professorship.
Stanley Osher is partially funded by STROBE NSF STC887
DMR 1548924,  NSF 00065369, and ONR N00014-20-1-2787.

\bibliography{references}
\bibliographystyle{plain}

\appendix

\section{Dataset Details}\label{sec:dataset_details}
We list below the parametric equations and the corresponding paramters used to generated the data. For ordinary differential equations (ODEs), we use SciPy's \texttt{solve\_ivp} function to compute solutions. All ODEs solutions are generated for $t\in[0,5]$. Partial differential equations (PDEs) and conservation laws are generated following the methodology described in \cite{sun2024towards}, to which we refer for detailed information on specific solvers. All PDEs and conservation laws solutions are generated for $(t,x)\in [0,5]\times [0,2]$ except otherwise stated below. To create the dataset, we sample 100 parameter values in the range $[Q-0.1Q, Q+0.1Q]$ for each of the 52 parametric equations, resulting in a total of 5200 unique equations. For the ODE set we sample 50 initial conditions per equation, while for the PDE set we sample 100 initial conditions per equation.  Additionally, for each of the equations, we generate 50 text descriptions using GPT-4. These descriptions focus on a variety of properties of the dynamics. In particular, some focus on scientific properties of the equation and its solution, such as order, linearity, main equation terms, or the presence of shocks or rarefactions in the solution. Others describe classical numerical methods which may be used to produce an approximate solution. Other focus on the natural dynamics that the equation describes, for example, the Lotka Volterra system describes the interactions between prey and predators in an ecosystem.
Below is a complete list of equations with the corresponding parameters of interest.
\bigskip

\textbf{ODE 1 (index 1):} 
\[
\frac{du}{dt} = a \sin(2 \pi t) u, \qquad a = 1
\]

\textbf{ODE 2 (index 2):} 
\[
\frac{du}{dt} = a e^{-t} + b \qquad [a,b] = [1,2]
\]

\textbf{ODE 3 (index 3):} 
\[
\frac{du}{dt} = a t^2 \cos(u) + b u \qquad [a,b] = [1,3/10]
\]

\textbf{ODE 4 (index 4):} 
\[
\frac{du}{dt} = a \sin\left( e^{-0.5 t} \sin(3 t) \right) + b u \qquad [a,b] = [2, 1/2]
\]

\textbf{ODE 5 (index 5):} 
\[
\frac{du}{dt} = a t \sin(u) \qquad a = 3/2
\]

\textbf{SIR System (index 6):} 
\[
\begin{cases}
\frac{dS}{dt} = -\beta S I \\
\frac{dI}{dt} = \beta S I - \gamma I \\
\frac{dR}{dt} = \gamma I
\end{cases} \qquad [\beta, \gamma] = [0.3,0.1]
\]

\textbf{Neural Dynamics System (index 7):} 
\[
\begin{cases}
\frac{dE}{dt} = \alpha E - \beta E I - \gamma E + 0.01 \sin(t) \\
\frac{dI}{dt} = \delta E - \epsilon I \\
\frac{dH}{dt} = \theta I - \phi H
\end{cases} \qquad [\alpha, \beta, \gamma, \delta] = [0.2 ,0.1 ,0.05 , 0.5]
\]

\textbf{Van der Pol Oscillator (index 8):} 
\[
\begin{cases}
\frac{dx}{dt} = y \\
\frac{dy}{dt} = \mu (1 - x^2) y - x
\end{cases} \qquad \mu = 2
\]

\textbf{Lotka-Volterra System (index 9):} 
\[
\begin{cases}
\frac{dx}{dt} = \alpha x - \beta x y \\
\frac{dy}{dt} = \delta x y - \gamma y
\end{cases} [\alpha, \beta, \gamma, \delta] = [1.5, 1, 3, 1]
\]

\textbf{FitzHugh-Nagumo System (index 10):} 
\[
\begin{cases}
\frac{dv}{dt} = v - \frac{v^3}{3} - w + I \\
\frac{dw}{dt} = \frac{1}{\tau} \left( v + a - b w \right)
\end{cases} \quad [I, a, b, \tau] = [0, 0.7, 0.8, 0.8]
\]

\textbf{Brusselator System (index 11):} 
\[
\begin{cases}
\frac{dx}{dt} = A + x^2 y - (B + 1) x \\
\frac{dy}{dt} = B x - x^2 y
\end{cases} \qquad [A,B] = [2,4]
\]

\textbf{Duffing System (index 12):} 
\[
\begin{cases}
\frac{dx}{dt} = y \\
\frac{dy}{dt} = -\delta y - \alpha x - \beta x^3
\end{cases} \qquad [\alpha, \beta, \delta] = [1, 0.2, 0.3]
\]

\textbf{Heat Equation (index 13):} 
\[
\frac{\partial u}{\partial t} = c_1 \frac{\partial^2 u}{\partial x^2} \qquad c_1 = 3\times 10^{-3}
\] 

\textbf{Porous Medium Equation (index 14):} 
\[
\frac{\partial u}{\partial t} = \frac{\partial^2}{\partial x^2} \left( u^m \right) \qquad m = 2,3,4, \; t_{final} = 0.1
\]

\textbf{Klein-Gordon Equation (index 15):} 
\[
\frac{\partial^2 u}{\partial t^2} + q_2^2 q_1^4 u = q_1^2 \frac{\partial^2 u}{\partial x^2} \qquad [q_1, q_2] = [1, 0.1], \; t_{final} = 1
\]

\textbf{Sine-Gordon Equation (index 16):} 
\[
\frac{\partial^2 u}{\partial t^2} + q \sin(u) = \frac{\partial^2 u}{\partial x^2} \qquad q = 1, \; t_{final} = 1
\]

\textbf{Cahn-Hilliard Equation (index 17):} 
\[
\frac{\partial^2 u}{\partial t^2} + q^2 \frac{\partial^4 u}{\partial x^4} + 6 \left( u \frac{\partial u}{\partial x} \right)_{x} = 0, \qquad q = 0.01, \; t_{final} = 0.5
\]

\textbf{Korteweg-De Vries (KdV) Equation (index 18):} 
\[
\frac{\partial u}{\partial t} + q^2 \frac{\partial^3 u}{\partial x^3} + u \frac{\partial u}{\partial x} = 0 \qquad q = 0.022, \; t_{final} = 1
\]

\textbf{Advection Equation (index 19):} 
\[
\frac{\partial u}{\partial t} + q \frac{\partial u}{\partial x} = 0 \qquad q = 0.5
\]

\textbf{Wave Equation (index 20):} 
\[
\frac{\partial^2 u}{\partial t^2} = q \frac{\partial^2 u}{\partial x^2} \qquad q = 0.5, \; t_{final} = 1
\]

\textbf{Reaction-Diffusion Equation Logistic (index 21):} 
\[
\frac{\partial u}{\partial t} = q_1 \frac{\partial^2 u}{\partial x^2} + q_2 R(u), \quad R(u) = u(1 - u) \qquad [q_1,q_2] = [3\times 10^{-3}, 1]
\]

\textbf{Reaction-Diffusion Equation Linear (index 22):} 
\[
\frac{\partial u}{\partial t} = q_1 \frac{\partial^2 u}{\partial x^2} + q_2 R(u), \quad R(u) = u \qquad [q_1,q_2] = [3\times 10^{-3}, 0.1]
\]

\textbf{Reaction-Diffusion Equation Bistable (index 23):} 
\[
\frac{\partial u}{\partial t} = q_1 \frac{\partial^2 u}{\partial x^2} + q_2 R(u), \quad R(u) = u^2(1 - u) \qquad[q_1,q_2] = [3\times 10^{-3}, 1]
\]

\textbf{Reaction-Diffusion Equation Square Logistic (index 24):} 
\[
\frac{\partial u}{\partial t} = q_1 \frac{\partial^2 u}{\partial x^2} + q_2 R(u), \quad R(u) = u^2(1 - u)^2 \qquad[q_1,q_2] = [3\times 10^{-3}, 1]
\]

\textbf{Burgers' Equation (index 25):} 
\[
\frac{\partial u}{\partial t} = -q_1 \left( f(u) \right)_x + \frac{q_2}{\pi} \frac{\partial^2 u}{\partial x^2}, \quad f(u) = \frac{1}{2} u^2  \qquad [q_1,q_2] = [1,0.01]
\]

\textbf{Inviscid Burgers (index 26):} 
\[
\frac{\partial u}{\partial t} = -k \left( f(u) \right)_x, \quad f(u) = \frac{1}{2} u^2 \qquad k = 1
\] 

\textbf{Conservation law Linear Flux (index 27):} 
\[
\frac{\partial u}{\partial t} = -q_1 \left( f(u) \right)_x + \frac{q_2}{\pi} \frac{\partial^2 u}{\partial x^2}, \quad f(u) = u \qquad [q_1,q_2] = [1,0.01]
\]

\textbf{Conservation law Cubic Flux (index 28):} 
\[
\frac{\partial u}{\partial t} = -q_1 \left( f(u) \right)_x + \frac{q_2}{\pi} \frac{\partial^2 u}{\partial x^2}, \quad f(u) = \frac{1}{3} u^3 \qquad [q_1,q_2] = [1,0.01]
\]

\textbf{Inviscid Conservation law Cubic Flux (index 29):} 
\[
\frac{\partial u}{\partial t} = -k \left( f(u) \right)_x + \frac{\partial^2 u}{\partial x^2}, \quad f(u) = \frac{1}{3} u^3 \qquad k =1
\]

\textbf{Conservation law Sine Flux (index 30):} 
\[
\frac{\partial u}{\partial t} = -q_1 \left( f(u) \right)_x + \frac{q_2}{\pi} \frac{\partial^2 u}{\partial x^2}, \quad f(u) = \sin(u) \qquad [q_1,q_2] = [1,0.01]
\]

\textbf{Inviscid Conservation law Sine Flux (index 31):} 
\[
\frac{\partial u}{\partial t} = -k \left( f(u) \right)_x + \frac{\partial^2 u}{\partial x^2}, \quad f(u) = \sin(u) \qquad k = 1
\]

\textbf{Conservation law Cosine Flux (index 32):} 
\[
\frac{\partial u}{\partial t} = -q_1 \left( f(u) \right)_x + \frac{q_2}{\pi} \frac{\partial^2 u}{\partial x^2}, \quad f(u) = \cos(u) \qquad [q_1,q_2] = [1,0.01]
\]

\textbf{Inviscid Conservation law Cosine Flux (index 33):} 
\[
\frac{\partial u}{\partial t} = -k \left( f(u) \right)_x + \frac{\partial^2 u}{\partial x^2}, \quad f(u) = \cos(u) k = 1
\]

\textbf{Fokker-Plank (index 34):} 
\[
\frac{\partial u}{\partial t} = D \frac{\partial^2 u}{\partial x^2} - \frac{D}{k_BT} \left( \nabla U(x)u \right)_x
\]
where $D = \frac{k_BT}{\gamma}$ where $k_B \approx 1.380649\times 10^{-23}$ is the Boltzmann constant, $T=300$ is absolute temperature and $\gamma = 6\pi\eta r$ represents the drag coefficient. $r = 0.1\times 10^{-6}$. The parameter of interest (randomized) is the fluid viscosity $\eta = 10^{-3}$. The potential is defined as $U(x) = 5\times 10^{-21} \cos\left(\frac{x}{0.1\times 10^{-6}}\right)$, and the values are set to $t_{final} = 0.1, x_{final}= 2\times 10^{-6}$.

\end{document}